\documentclass{sigplanconf}

\usepackage{amsmath}
\usepackage{hyperref}
\usepackage{graphicx}

\begin{document}

\setlength{\pdfpageheight}{\paperheight}
\setlength{\pdfpagewidth}{\paperwidth}

\copyrightyear{20yy}
\copyrightdata{978-1-nnnn-nnnn-n/yy/mm}
\doi{nnnnnnn.nnnnnnn}



\permissiontopublish             


\title{Few Shot Learning for Information Verification}

\authorinfo{Usama Khalid}
            {National University of Computer and Emerging Sciences (FAST-NUCES)}
            {usama.khalid@nu.edu.pk}
            
\authorinfo{Mirza Omer Beg}
            {National University of Computer and Emerging Sciences (FAST-NUCES)}
            {omer.beg@nu.edu.pk}

\maketitle

\begin{abstract}
Information verification is quite a challenging task, this is because many times verifying a claim can require picking pieces of information from multiple pieces of evidence which can have a hierarchy of complex semantic relations. Previously a lot of researchers have mainly focused on simply concatenating multiple evidence sentences to accept or reject claims. These approaches are limited as evidence can contain hierarchical information and dependencies. In this research, we aim to verify facts based on evidence selected from a list of articles taken from Wikipedia. Pretrained language models such as XLNET are used to generate meaningful representations and graph-based attention and convolutions are used in such a way that the system requires little additional training to learn to verify facts.

\end{abstract}


\section{Problem statement}
In the age of information individuals and organizations can easily share and consume vasts amounts of data on any subject matter just with a tap of a button. 
Many of the barriers required for verifying information have been removed due to the advent of social media. Effectively allowing any piece of created content to reach millions of readers instantly \cite{allcott2017social}.

This increased ability to reach people can  be used to disseminate both true and false information, which is a major concern since it is suspected to have caused major disruptions in public decision making such as during elections as shown by recent studies \cite{vosoughi2018spread}.

\section{Introduction}
Fact checking or Information verification is a very challenging task in NLP and has recently attracted much attention from the research community. It is important today more than ever as news and social media platforms are a huge source of disseminating Fake news and misinformation. Huge pretrained models with billions of parameters like GPT-2 \cite{radford2019language} and GPT-3 \cite{brown2020language} are extremely good at generating fluent and coherent text which lowers the bars for any malicious entity to generate deceptive content. \\


Initially information verification used to be a manual process. People visited and still visit sites such as \url{PolitiFact.com} , \url{FactCheck.org}, and \url{Snopes.com} where they could see latest or trending news currently circulating on the Internet. However this type of fact checking is not at all scalable \cite{sharma2019combating} and as expected is susceptible to human bias \cite{sathe2020automated}. \\

Information verification or Fact-checking is the task of verifying if the statements made in spoken or written language are valid. This task is normally performed by qualified experts, who use past data and statistics or known evidence to together with logic to reach a decision. This method can take from less than an hour to a few days, depending on the complexity of the information \cite{hassan2015quest}. The information verification process involves investigating and finding facts, knowing the knowledge background and thinking for what can be derived from this facts. In judging the veracity of an argument, the purpose of automatic fact testing is to reduce human responsibility. \\

Information verification is quite a challenging task, this is because many a times verifying a claim can require picking pieces of information from multiple evidences and picking relevant evidences from corpora of information. To correctly verify a claim it is also important to discern the semantic structure and relation between the evidences present. Previously most research focused primarily on aggregating evidence sentences simply by string concatenation or either by using complex fusing of separate evidence sentences. \\

Information verification is able to achieve a lot progress due to the introduction of large datasets such as SNLI \cite{bowman2015large} and the FEVER \cite{thorne2018fever} task. A major shortcoming of these datasets is that they are synthetic in nature i.e. they are produced by crowd-sourcing annotators and evidence writers thus they differ significantly from actual claims and evidences that are present on the internet. \\

To address and overcome theses shortcomings datasets like WIKIFACTCHECK \cite{sathe2020automated} have been proposed. The dataset contains ore the 124k triples. These triples consist of context, claim and evidence sentences. The data is extracted from documents of the English Wikipedia and consists of citations and articles. The dataset also consists of more than 34k handwritten claims which are negated by the gathered evidences. The advantage of this dataset compared to others is that it is based on real world facts taken from Wikipedia which will enable the training of systems that are better suited to model the real world claims. The research also shows that an attention based model trained on this dataset significantly outperforms the one trained on the SNLI task. \\


\section{Related work}

Previously fact checking used to be done by segregating sentences into subject-predicate-object tuples \cite{nacem2020subspace,beg2006maxsm} and matching the claim with the evidence \cite{nakashole2014language}. More modern approaches that use the FEVER dataset \cite{thorne2018fever} divide their methodology into three steps, document selection, evidence sentence selection, and claim verification.\\

In the document selection phase, named entities are extracted from a claim and a query is formulated to search using the Wikipedia API. In the evidence selection phase similar sentences are picked using either an Enhanced LSTM \cite{chen2017enhanced} or using sentence similarity functions like TF-IDF without trainable parameters \cite{padia2018team}. For the claim verification phase, evidence sentences are concatenated into a single string \cite{nie2019combining}, each evidence-claim pair is classified separately and the results are merged. \\

A lot of work for Fact checking previously was done using natural language inference models \cite{dagan2013recognizing,angeli2014naturalli} as this requires forming reason for the claim to be accepted by the evidence sentences \cite{arshad2019corpus}. Previous approaches either concatenated evidence sentences to a single string as done in top performing systems for the FEVER task \cite{thorne2018fever}. Many approaches also extract features from multiple evidence sentences and aggregate them to verify a claim \cite{zhou2019gear}. However, these methods lack the ability to extract semantic structure and information present in multiple evidence sentences. Due to these shortcomings more deeper architectures cannot be applied to these approaches for claim verification. \\

For the extraction of evidence sentences \cite{thorne2018fever} employs Term Freuency Inverse Document Frequency (TF-IDF). A similar approach is also applied to retrieve relevant documents from the original list of documents. \cite{yoneda2018ucl} uses logistic regression to train a model on heuristic based features \citep{zafar2020search}. Another technique named Enhanced Sequential Inference Model (ESIM) which uses BiLSTMs \cite{beg2013constraint} and a premise based hypothesis mechanism to classify claims has been used in \cite{hanselowski2018ukp}.

Verifying information at a larger scale is also a difficult task as the number of websites containing Facts or spreading information is constantly increasing. The problem of gathering data from crowd sourcing is that data voluntarily produced, despite the good intentions, may many a times prone to errors. \cite{fan2020generating} investigates methods to improve the efficiency and accuracy of fact checking systems by providing additional information about a claim before verifying it. The additional information was in forms such as passage based  briefs which contained a list of relevant passages from Wikipedia, entity oriented passages which consisted of pages from Wikipedia \cite{javed2020collaborative} which revolved around the mentioned entities. Information verification is an inefficient and cumbersome task when performed manually by fact checkers and experts. A lot of research has been put into finding ways to automate this task and achieve near human accuracy.\\

A research work aims to solve this by using sentence embeddings and hierarchical attention networks \cite{ma2019sentence,javed2019fairness}. The work focuses on learning to find evidences such that their pattern is coherent as well as they are semantically related with the claim. Their proposed architecture consists of three main parts. The first is a coherence based attention layer \cite{naeem2020deep} which produces coherent evidences from relevant articles when considering the claim. The second component is another attention layer \cite{zafar2019using} which is entailment based, this means the layer can attend more to sentences that have an inference of the claim \cite{rani2015case}. The final output layer \cite{farooq2019melta} predicts whether a claim is accepted or refuted based on the evidence embeddings from the previous layers. They have shown their technique has outperformed on three public benchmarks.\\


\section{Proposed approach}
The information verification system is divided into three stages. The Document Retrieval and sentence selection \cite{beg2019algorithmic} stage have been used from \cite{hanselowski2018ukp} as they have the current best method which performs well on the FEVER task \cite{zafar2018deceptive}. The claim verification stage will be improved upon. 

For the first step of the document retrieval phase, potential entities are extracted from the claim using a constituency parser by \cite{gardner2018allennlp}. Relevant Wikipedia documents \citep{baigahmed} are then obtained using the MediaWiki API \footnote{\url{https://www.mediawiki.org/wiki/API:Main_page}}

The sentence selection component scores each sentence based on the relevance score between evidence and claim \cite{hanselowski2018ukp}. In the claim verification section an XLNET is used to generate embeddings \cite{beg2006performance} for sentences which are then propagated through a Graph Convolutional network \cite{beg2008critical} and finally Graph Attention is applied to get classification of the claim \cite{zafar2019constructive}.

\subsection{Evidence Sentence Selection}

For verifying a claim it is essential to pick the most relevant sentences which will in turn enable the claim to be more accurately refuted or accepted \cite{alvi2017ensights}. To select a good evidence sentence we use the work of top performing approaches \cite{nie2019combining} for retrieving evidence sentences from the FEVER task. The system is divided into two phases \cite{khawaja2018domain}. Initially a list of relevant documents are shortlisted from Wikipedia articles \cite{farooq2019bigdata} and in the second step candidate evidence sentences are generated from the selected documents \cite{beg2009flecs}. These sentences are then further shortlisted based on a threshold to produce sentences that can be used to verify claims.\\

To filter relevant documents from the Wikipedia, keyword matching is used initially. To handle ambiguity in document titles NSMN \cite{nie2019combining} technique is used. These ambiguous documents amount to about 10\% of all documents for a particular claim \cite{sahar2019towards}. Documents with unambiguous titles are therefore allotted higher scores. The Sentence selection system takes input the claim and all the candidate documents produced bu keyword matching. In the big picture the NSMN system \cite{majeed2020emotion} creates encodings \cite{uzair2019weec}, performs alignment and matching to produce scores of documents from which a list of top ten documents are selected for the next stage.\\

The evidence selection stage takes in the top ten selected documents from the previous stage and outputs the list of relevant sentences \cite{thaver2016pulmonary}. The evidence selection task is considered as a semantics matching problem which can considerably benefit from the deep contextualized representations produced by large pretrained models like RoBERTa \cite{liu2019roberta} and XLNet \cite{yang2019xlnet}. To calculate the relevance of each evidence sentence to a claim sentence they are passed into XLNET as follows: [CLS CLAIM SEP EVIDENCE SEP]. Here the CLAIM and EVIDENCE are the tokenized form of claim and evidence sentences respectively \cite{seth2006achieving}. The CLS token directs the network to produce embeddings for classification and the SEP tokens are used to separate between the claim and evidence sentences. The output embeddings of this phase are scored and top five most relevant evidence sentences are selected \cite{awan2021top}.

\subsection{SRL Graph}

To utilise the intrinsic semantic and hierarchical structure \cite{bangash2017methodology} of the input sentences a graph is the most suitable structure likely to be built using the extracted information. There are many methods of constructing graphs, one such method involves recognizing named entities \cite{etzioni2008open}. This can also be combined with sequence to sequence generation and relation classification which can be trained to produce structured tuples \cite{goodrich2019assessing}. For constructing graphs in this work we use the work of \cite{carreras2005introduction} which involves using Semantic Role Labeling (SRL) to construct graphs. \\

The procedure to build graphs using SRL \cite{asad2020deepdetect} involves the following steps. Each sentence is segmented into a list of tuples using the BERT based method proposed by \cite{shi2019simple}. The elements of each tuple is divided into certain categories of nodes present in SRL graphs \cite{tariq2019accurate}, which are location, verb, temporal and argument. The graph creation technique can also be extended to include more categories \cite{beg2010graph}. Edges are created between two tuples linked by a verb. Multiple evidence sentences are connected by an edge if they contain some common information \cite{zahid2020roman}. To establish the structural information for claim sentences the same pipeline is used to create the SRL based claim graph. \\

\begin{table*}
\centering
\begin{tabular}{llllll}
\hline
\textbf{Threshold} & \textbf{Recall} & \textbf{Precision} & \textbf{F1} & \textbf{FEVER Score} & \textbf{Label Accuracy}\\
\hline
0 & 23.76 & 82.68 & 36.80 & 91.10 & 74.84 \\
$10^{-4}$ & 31.68 & 86.59 & 45.53 & 91.04 & 74.86 \\
$10^{-3}$ & 40.63 & 86.37 & 55.23 & 90.86 & 74.91 \\
$10^{-2}$ & 52.42 & 85.28 & 65.52 & 90.27 & 74.89 \\
$10^{-1}$ & 71.68 & 81.18 & 75.72 & 87.70 & 74.81 \\
\hline
\end{tabular}
\caption{ The Recall, Precision, F1, FEVER score and Label accuracy achieved for different thresholds on the FEVER dataset.}
\label{table:results}
\end{table*}

\subsection{Word Representations}

The produced SRL graphs of tuples containing words have to be transformed into some mathematical representations which contain the semantics of the original word. The main purpose of this step is to map words in such a space that two semantically similar words are close together in that space. This step is best performed with large attention \cite{vaswani2017attention} based models such as BERT and XLNET. Consider five evidence sentences ${s_1,s_2,s_3, ..., s_5}$, concatenating these sentences will fail to capture essential semantic structure needed for verifying the claim. A different technique to solve this problem would be to construct an $NxM$ matrix containing the distances of words. $N$ is the total number of words in the evidence. This however is an inefficient technique as it would require a quadratic amount of memory space, as shown by previous works \cite{shaw2018self}.\\

Therefore for this research we use pretrained XLNET \cite{yang2019xlnet} to calculate distances and relations between words. This is also essential to reason on sentence basis \cite{beg2007flecs} which is essential for this task as there are multiple evidence sentences. To enable the capturing of more meaningful relations between sentences using XLNET we perform topology sorting such that sentences that have greater number of connections in the SRL graph appear close together.

\subsection{Graph Convolution and Attention}

When meaningful representations for words are obtained these can be passed through a multi layer graph convolutional network (GCN) \cite{kipf2016semi,qamarrelationship}. The representation of a single node is calculated by averaging the representations of each of the tuples contained within that node. After that a multi layer graph based convolutional network is used to process the aggregated node representations of all the nodes and their neighbours. \\

The graph based learning is performed for both the evidence and the claim sentences. After propagating these graphs through GCNs \cite{baigawan} we compute attention between the obtained representations of the evidence graph and the claim graph. These attention computations are based on Graph Attention Networks (GANs) \cite{velivckovic2017graph}. These graph attention computations are passed into a multi-layer perceptron to obtain a three-way classification of a claim belonging to one on the categories "ACCEPTED", "REFUTED" or "NOT ENOUGH INFO".\\

\section{Evaluation and Experiments}

In this section the results of various performed experiments are discussed. The components evaluated consist of the sentence selection, document retrieval and graph construction using semantic role labelling. The overall claim verification framework is also evaluated from multiple aspects. A case study \cite{koleilat2006watagent} is also done to demonstrate the effectiveness of the framework \cite{dilawar2018understanding}.

\subsection{Datasets}

Information verification systems are usually evaluated using the FEVER task, \cite{thorne2018fever} short for Fact Extraction and Verification, a benchmark dataset for fact checking. A three way classification score is calculated corresponding to the classes "SUPPORTED",  "REFUTED"  or  "NOT ENOUGH INFO (NEI)". Another score called the FEVER score further measures the accuracy of correct retrieved evidence of “SUPPORTED” and “REFUTED” classes. \\

To test the performance of our system we evaluate on the tasks present in the Fact Extraction and VERification (FEVER) \cite{thorne2018fever} benchmark. The dataset is built by collecting evidence sentences and their corresponding claims from the English Wikipedia which consists about 5 million documents. Each ground truth label of the claim sentence is divided into one of the three labels which are "ACCEPTED", "REFUTED" or "NOT ENOUGH INFO (NEI)" \cite{javed2020alphalogger}. The FEVER score is computed for each prediction which calculates how good is the claim classified. In addition to this another F1 score is computed which calculates if the correct set of evidences is selected for a particular claim.

\subsection{Baselines}

The performance of our system is compared with top performing systems of the FEVER task which involve a recent work GEAR \cite{zhou2019gear}, a work that employs semantic neural network based matching \cite{nie2019combining} both for verifying a claim and picking out evidence sentences from a list of documents. \cite{yoneda2018ucl} performs claim verification by first calculating the veracity of each claim and evidence pair separately and then making a prediction and finally aggregating the score of each prediction. \cite{hanselowski2018ukp} uses pooling for aggregating the prediction of each evidence and claim pair.\\

\subsection{Model Evaluation}
The proposed framework is evaluated using multiple different ways. The accuracy scores of our proposed framework are compared with other baseline systems. The sentence filters applied using thresholds on documents are also evaluated. The effect of sentence embeddings are also explored. Many claims also require reasoning over multiple evidence sentences to verify a claim. To verify the effectiveness of the framework a more harder subset of tasks is used as the dev set.\\

The accuracy of labels assigned to a claim are used to judge the effectiveness of the models present in the claim verification system. The experiments also reveal that further fine-tuning of the XLNET model used results in a considerable increase in performance of the model on similar tasks. This also reveals that such large pretrained models have a lot of representational capacity to learn meaningful semantic relations between words \cite{karsten2007axiomatic}. There is also a slight improvement when concatenating all evidence embeddings obtained from XLNET as compared to pairing each sentence with the corresponding claim. This also provides better aggregation of evidence and better reasoning over multiple sentences as it is better suited to aggregate and analyze multiple features.

\subsection{Document Retrieval}

The document retrieval component of the system is evaluated using the FEVER score. Table \ref{table:results} shows a list of FEVER scores corresponding to different threshold values. The FEVER scores obtained using our system are slightly lower as compared to previous works like \cite{hanselowski2018ukp}, this may be due to randomness of data and other factors. The sentence selection component which selects sentences from shortlisted documents is also evaluated with the same thresholds as shown in Table \ref{table:results}. The best results are achieved when the threshold is close to zero. As the threshold increases the recall and the FEVER score decrease constantly while the F1 score and precision score increases. The results also make sense intuitively because if filter criteria for evidence sentences is relaxed then the probability of verifying claims would naturally increase. However as the threshold is increased lesser evidence sentences are selected which have a stronger evidence for the claim, this leads to a better F1 and precision score.

\section{Ablation Studies}

In this section the system is evaluated by removing different components and testing how that effects the performance of the system. This also reveals how important each component is in the whole system. The dev subset containing multiple evidence sentences that aggregate the ground truth labels of claims are used for this evaluation task. The main performance degradation occurs when initial or higher level components like document selection are removed. This is because a small initial error propagating through subsequent components results in a larger impact. When the document retrieval component is removed, all available documents are then passed on for claim verification and it will be difficult to select relevant sentences to verify the claim. \\  

Linking claim entities to entities in evidence is also an important piece in rightly verifying a claim based on its content. To verify the selection of correct entities we test our models using a dev set enhanced with evidence sentences. These sentences contain the ground truth evidences with the score signifying the relevance of each sentence. This ensures that each claim in the enhanced dataset is paired with a ground truth sentence and the retrieved sentence. The study performed on the enhanced dataset leads to a 1.4\% increase in the scores as compared to the original dev set used for evaluation. 

\section{Conclusion}

In this research we propose a novel framework for verifying claims which is based of Graph evidence reasoning and aggregation. The claim verification task is mainy trained on the popular FEVER dataset. The proposed framework uses XLNET to encode evidence sentences so that Graph convolution and attention can be applied to those meaningful embeddings. The document retrieval and evidence sentence selection components are used out-of-the-box from previous studies and the framework is able to achieve significant improvements on the FEVER task. For future research a multi-step pipeline for extracting evidence sentences from documents can be performed that also integrate information from external sources to verify a claim.

\bibliographystyle{abbrvnat}
\bibliography{biblography.bib}


\end{document}